# Multimodal Biometric Authentication Using Choquet Integral and Genetic Algorithm

Anouar Ben Khalifa, Sami Gazzah, Najoua Essoukri BenAmara

*Abstract*—The Choquet integral is a tool for the information fusion that is very effective in the case where fuzzy measures associated with it are well chosen. In this paper, we propose a new approach for calculating fuzzy measures associated with the Choquet integral in a context of data fusion in multimodal biometrics. The proposed approach is based on genetic algorithms. It has been validated in two databases: the first base is relative to synthetic scores and the second one is biometrically relating to the face, fingerprint and palmprint. The results achieved attest the robustness of the proposed approach.

*Keywords*—Multimodal biometrics, data fusion, Choquet integral, fuzzy measures, genetic algorithm.

## I. INTRODUCTION

THE Choquet Integral (CI) is a tool for the information fusion which can generalize many operators such as the Ordered Weighted Averaging, the arithmetic sum, the minimum, the maximum… It has been employed as an aggregation tool to calculate a global score, taking into account the magnitudes of criteria expressed by a fuzzy measure, in various applications such as: the regulation of multimodal transport systems, the fusion of information, the recognition of graphic symbols, the management of human and material resources, the air traffic control [26]… Indeed, in [18], the authors proposed two approaches for the biometric face identification. In the first approach, they cut the image into three zones (eyes, nose and mouth) to construct a multimodal system at the base of these zones. In the second approach, the face underwent a wavelet decomposition to obtain four matrices that corresponded to the approximation matrix and the vertical, diagonal and horizontal details, thus, four unimodal systems were constructed by this transformation. In both approaches, the fusion was carried through the CI with fuzzy measures given by the classification rate. In [19], the authors used the CI in the field of the classification of acoustic events. They proposed the fusion of multi-source acoustic information and they claimed that the fusion through the CI is more common when the fusion of features is delicate. The adopted fuzzy measures were calculated through entropy. In [26], the authors proposed a decision support system to the regulators of multimodal transportation. This proposal comes in response to the needs of regulators to be assisted in their decision-making face to random perturbations that affect the multimodal network. To model the transport system, the regulation problem has been reduced to a problem of decision fusion. The fusion tool used is the CI takes account of possible interactions between the different criteria involved in decision making. The results achieved show that the use of the CI is a promising approach. Table I presents a selection of the work performed with the CI.

The exploration work on the fusion of information by the CI shows the diversity and the multitude of these application fields. Indeed, we can find it in all the areas requiring the aggregation of information. Its efficacy to take into account interactions between different sources of information makes it a robust tool for fusion. Nevertheless, the use of the CI is not trivial; the restriction resides in the choice of fuzzy measures. Indeed, the difficulty that has slowed the exploitation of the CI is the choice of the most appropriate fuzzy measures, since we need to define a measure that has real meaning for each source of information. Several methods for determining fuzzy measures have been proposed in the literature. These methods are based on an expert election, a statistical analysis or an optimization and they generally depend on the application domain [25], [26]. Far as we know, there is not a generic technique for determining fuzzy measures operating effectively on any problem of information fusion. This leaves the field open to some expertise.

In [1], we proposed a biometric verification of identity based on: the face, the off-line signature and the off-line handwriting. The fusion of the three biometric modalities was operated by the CI with fuzzy measures given by the confusion matrix and entropy. In this paper, we propose to exploit the CI for data fusion in multimodal biometrics. In order to demonstrate the transparency of the CI towards the biometric modalities, we present other system based on biometric modalities: face, fingerprint and palmprint. We are particularly interested in the fusion step of the three modalities. Thus, we present a new approach to the score-level fusion by means of the CI and the Genetic Algorithms (GA). Given its explorer and exploiter character of the space of solutions, we have focused our choice on the GA to calculate the most appropriate fuzzy measures for our fusion problem.

In the following, we give an overview on data fusion in multimodal biometrics. In Section III, we present the three unimodal systems. In Section IV, we introduce the CI and the GA. In Section V, we present the proposed fusion approach. The experiments and the results achieved are clarified in

Anouar Ben Khalifa is with the Research Unit of Advanced Systems in Electrical Engineering. University of Sousse - Tunisia (phone: +216 73 369 500; fax: +216 73 369 506; ben_khalifa_anouar@ yahoo.fr).

Sami Gazzah is with the Research Unit of Advanced Systems in Electrical Engineering. University of Sousse - Tunisia (phone: +216 73 369 500; fax: +216 73 369 506).

Najoua Essoukri Ben Amara is with the Research Unit of Advanced Systems in Electrical Engineering. University of Sousse - Tunisia (phone: +216 73 369 500; fax: +216 73 369 506).

Section VI.

TABLE I
A SELECTION OF THE WORK PERFORMED WITH THE CHOQUET INTEGRAL

| Ref. | Fields | Fuzzy measures | Performances (%) |
|---|---|---|---|
| [20] | Multimodal gesture recognition: fusing information from camera and 3D accelerometer data. | Similarity measure between the camera and accelerometer modules. | Camera only : RR = 76.7<br>Accelerometer only : RR = 70<br>Fusion (CI) : RR = 92.7 |
| [21] | Multi-biometric authentication system: fusing data from face and voice. | Classification rate for each unimodal system. | Face only : RR = 96.62<br>Voice only : RR = 97.7<br>Fusion (Sum) : RR = 99.7<br>Fusion (CI) : RR = 99.99 |
| [23] | Fusion of multiple Support Vector Machine classifiers. Application: *UCI data set* (iris, wine, glass and heart). | Measure based on confusion matrix. | Best result : Heart data<br>Best SVM : RR = 77.2<br>Majority vote fusion: RR=77.5<br>Fusion (CI) : RR = 80.9 |
| [24] | Combination of criteria for evaluating the overall satisfaction of patients in order to effectively manage a hospital. | • Entropy<br>• Complexity<br>• Cardinality | • Average correlation = 0.71<br>• Average correlation = 0.72<br>• Average correlation = 0.65 |
| [22] | Analysis of the quality of composite material: fusion of several attributes related to texture homogeneity and intensity gradient orientation extracted from X-ray images. | Entropy of the attribute images. The attributes are Gabor wavelet, Haar wavelet and gradient orientation variation. | Gabor wavelet : CR = 89<br>Haar wavelet : CR = 92<br>Gradient orientation : CR = 58<br>Fusion (Decision) : CR = 87<br>Fusion (CI) : CR = 95 |
| [25] | Multi criteria aid to the decision. Application: Marketing et benchmarking sites e-commerce. | Membership functions | -- |

(RR: Recognition Rate, CR: Classification Rate)

## II. DATA FUSION IN MULTIMODAL BIOMETRICS

Uni-modal Biometric systems have limitations which are generally due to noisy sensor data, non-universality and lack of individuality of the biometric trait, absence of an invariant representation for the biometric trait and susceptibility to circumvention [40]. All these limitations can be reduced by using multiple biometric modalities in the same system hence forming a multimodal biometric system. The combination of two or more modalities can be done at four different levels: at the signal level, the feature extraction level, the score level and the decision level [1], [2]. Current research is oriented towards determining the best level of fusion and the optimal fusion method [5], [6], [10], [13].

The fusion at signal level as well as the fusion at the feature level, used to combine data before they are distorted by analysis procedures and treatments, requires only one phase of learning for all modalities. However, this type of fusion is not widely used because it requires homogeneity between data. The fusion at the decision level is often used for its simplicity; it is a combination of binary decisions through operators such as majority voting, AND and OR… In [4], the author describes various fusion methods at the decision level. These methods are very simple but use very little information. The fusion at the score level is the most common type of fusion since it manipulates more information than the fusion at the decision level [3]. It can be applied to all multimodal biometric systems, with very effective methods, such as mean, product, minimum, maximum, weighted average and methods based on classifiers.

Many multimodal biometric systems have been proposed and compared in the literature. The state of the art about fusion in multimodal biometrics cannot conclude about the existence of a technical or a generic fusion level efficiently operating on any multimodal system. Nevertheless, the fusion at the score level is the most dominant combination into multimodal biometrics [14]; it has been widely studied in the literature and generally has shown its superiority over other levels of fusion. Indeed, in [11], the authors propose a multimodal system based on the finger-knuckle-print and the palmprint; they compare simple fusion methods at the score and at the decision level: *Sum, Weighted, Min, Max rules*. They conclude that the *Sum* is the best fusion technique and it surpasses the *mean* which gives a comparable performance to the best unimodal system. However, the fusion at the decision level gives poor results by getting a lower performance than the best unimodal system. *Wang* et *al.* in [13] found similar results on a database formed by *PolyU* and *CASIA* respectively for the palmprint and the iris. They compare fusion methods to the score level based on the *Gaussian mixture model*, *Sum* and *Prod*, with the fusion at the decision level based on *Max* and *Min*. The obtained results demonstrate once again the robustness of the fusion at the score level. The same conclusions were found in the works of Pigeon [12] on the M2VTS database. In this work, the author compares the fusion at the decision level (AND and OR) with the fusion at the score level (arithmetic average) and he argues that a fusion based on a combination of scores provides the best performance than that a fusion based on the grouping of individual decisions. In [10], Ferrer et al. propose a bimodal system based on face and lips. They demonstrate on two different databases (GPDS ULPGC-Face Database, the PIE Face Database) that the scores level fusion is more effective than by the concatenation of features.

Table II presents a selection of multimodal biometric authentication systems specifying for everyone the level and the fusion method used and the results achieved.

TABLE II
PREVIOUS WORK ON DATA FUSION INTO MULTIMODAL BIOMETRICS

| Ref | Biometrics | database | Fusion level | Performances (%) | |
|---|---|---|---|---|---|
| | | | | Unimodal system | Multimodal system |
| [5] | • Hand geometry<br>• Palm print<br>• Fingerprint. | 109 people | Feature level | • FA = 0.21<br>FR = 0.18. | FA = 0.1<br>FR = 0.4 |
| | | | Score level: Sum. | • FA = 0.01<br>FR = 0.25 | FA = 0.13<br>FR = 1.30 |
| | | | Decision level : majority voting | • FA = 0.01<br>FR = 0.20 | FA = 0<br>FR = 0.15 |
| [6] | • Face.<br>• ECG. | 35 people | Feature level | • RR = 91 | RR = 99 |
| | | | Score level : prod | | RR = 94 |
| | | | Decision level : voting fusion | • RR = 55 | RR = 66 |
| [8] | • Frontal face<br>• Gait silhouette | 70 people | SUM rule | • RR = 40 | RR = 71 |
| | | | Bayesian rule | | RR = 70 |
| | | | Confidence weighted score sum | | RR = 58 |
| | | | Rank sum | • RR = 39 | RR = 68 |
| [7] | • Face<br>Fingerprint | BANCA<br>50 people | Feature level | • RR = 88.9 | RR = 97.41 |
| | | | Score level : Sum | • RR = 91.82 | RR = 94.77 |
| [9] | • Fingerprint<br>• Finger-vein | 64 people | Feature level | • RR = 89.06 | RR = 99.68 |
| | | | Score level : Sum | • RR = 97.18 | RR = 98.75 |
| [10] | • Lips<br>• Face | GPDS-ULPGC<br>50 people | Feature level | • EER = 2.59 | EER = 2.32 |
| | | | Score level : Sum | | EER = 0.43 |
| | | | Prod | • EER = 3.48 | EER = 0.44 |

(FA: False Acceptance, FR: False Rejection, RR: Recognition Rate, EER: Equal Error Rate)

Table II confirms what we have introduced. Indeed, most research in multimodal biometrics has concentrated on the score-level fusion as it turns out to be more efficient than the rest of the fusion levels.

### III. THE PROPOSED UNIMODAL SYSTEMS

In this section, we present the three unimodal systems based respectively on the face from the Essex database [41], the palmprint and fingerprint of the PolyU database [37], [39]. The three biometric authentication systems respond to the traditional model of a system of pattern recognition. They consist of the following steps: acquisition, characterizing, learning and decision.

In our work, the characterization is based on a Discrete Wavelet Transformation (DWT). The Daubechies9 at level 2 of decomposition has been selected for the face and fingerprint while the Symlet6 at level 2 of decomposition has been used for the palm print modality. The features used for each modality are composed by the mean and standard deviation from an approximation image and the standard deviation of the vertical, horizontal and diagonal details. For learning, we have opted for a modular architecture based on the support vector machines with the RBF kernel. Fig. 1 shows a block diagram of the three unimodal systems.

The classification module is used during authentication to compare the reference characteristics and testing. Thus, each modality returns a similarity score relative to the person to be authenticated.

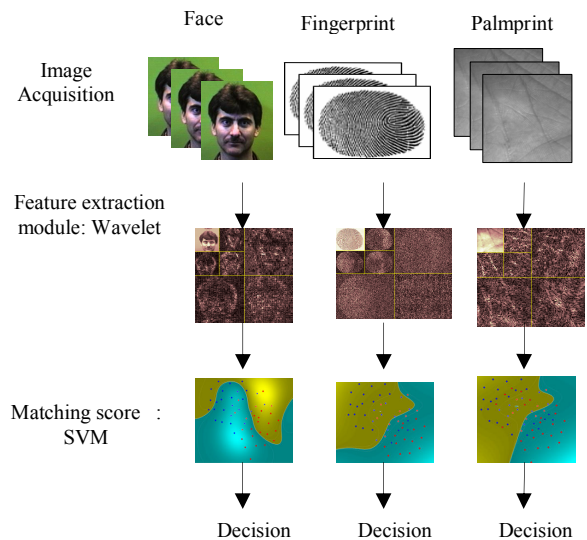

Fig. 1 A block diagram of the proposed unimodal biometric systems based on face, fingerprint and palmprint.

The scores are normalized between 0 and 1 with the *MinMax* method. However, we use the normalized scores between 0 and 1 where 0 indicates a complete rejection (presence of an impostor) and 1 indicates certain acceptance (presence of a client). In order to demonstrate the robustness of our approach, we compare in Table III the performance of three unimodal systems with the works in the literature.

In Table III, we find that the performance of the three unimodal systems differ from one approach to another. Different characterizations methods (geometric, global, local and hybrids) have been exploited, and various classification techniques, from simple Euclidean distance to hybridization of classifier, have been validated. These methods are complex in terms of the approach that we propose. Indeed, a textural analysis by the DWT and a modular architecture based on SVM has given good performances; i.e., an EER ranges from 6.5% for the fingerprint to 6.75% for the palmprint and 2.51% for face.

TABLE III
PREVIOUS WORK ON THE PROPOSED DATABASES

| Databases | Ref | Methods | Performances (%) |
|---|---|---|---|
| Exess face database [41] | [27] | • Statistical feature and Neural Network. | • RR = 98 |
| | | • Fast Fourier Transform and Neural Network. | • RR = 89 |
| | [28] | • Curvelet texture feature extraction and PCA. | • RR = 98.41 |
| | [29] | • Spatially Confined Non-negative Matrix Factorization. | • RR = 95.17 |
| | [30] | • Wavelet Transforms (DB7) and Zernike Moments. | • RR = 94.26 |
| | [31] | • Discrete Wavelet Transform and Structural HMM. | • RR = 90.7 |
| | [32] | • Hypercomplex Gabor Filter and Euclidean distance | • RR = 90.2 |
| | Proposed | • *Discrete Wavelet Transform and Modular SVM.* | • *RR = 97.49* |
| PolyU palm print database [39] | [13] | • Phase only Correlation function | • RR = 85 |
| | [33] | • Gabor transform, Wavelet transform and Neural Network | • RR = 95 |
| | [34] | • Gaborplam and kernel PCA and RBF classifier. | • RR = 65.99 |
| | [35] | • Discrete Wavelet Transform, Gabor filter and Euclidean distance. | • RR = 94.45 |
| | [36] | • Histogram equalization, Discrete cosine transform, mean square error | • RR = 92.05 |
| | Proposed | • *Discrete Wavelet Transform and Modular SVM.* | • *RR = 93.25* |
| PolyU HRF database [37] | [37] | • Texture information, neighboring minutiae and SVM. | • EER = 17.67 |
| | | • The spare representation technique and the weighted random sample consensus. | • EER = 6.59 |
| | [38] | • The correspondences between pores and the random sample consensus | • EER = 20.49 |
| | | • Minutia-based pore matching method | • EER = 30.45 |
| | Proposed | • *Discrete Wavelet Transform and Modular SVM.* | • *EER = 6.5* |

(RR: Recognition Rate, EER: Equal Error Rate)

In Fig. 2, we illustrate the variations of the EER relative to each modality for a selection of individuals. We find that the performances with the same person differ from one modality to another, which justifies the fusion and makes it interesting.

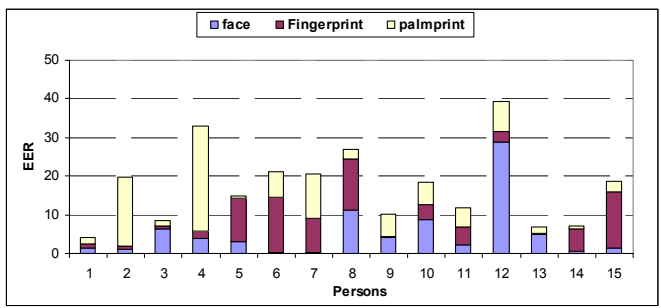

Fig. 2 Variation of the EER for a selection of individuals

## IV. THE BASIC CONCEPTS

In this section, we recall the basic concepts of the Choquet integral used as a tool for data fusion and genetic algorithms introduced for the calculation of fuzzy measures.

### A. Fuzzy Measure and Choquet Integral

*Fuzzy measure*: We call a fuzzy measure [15, 16] the function $m$: P(Y) → [0, 1] satisfying the conditions (1) and (2):

$$m(\emptyset) = 0, \ m(Y) = 1 \quad (1)$$

$$m(A) \leq m(B), \text{ if } A \subset B \text{ and } A, B \in P(Y) \quad (2)$$

$m(A)$ represents the importance or the power of coalition A for the fusion problem. Following this definition, Sugeno [17] introduced the fuzzy measure $m_\lambda$ which comes with an additional property:

$$m(A \cup B) = m(A) + m(B) + \lambda\, m(A)\, m(B). \quad (3)$$

For all (A, B ⊂ Y), (A ∩ B = ∅), and for $\lambda > -1$, $\lambda$ is determined by solving the following equation:

$$\lambda + 1 = \prod_{i=1}^{n}(1 + m^i) \quad (4)$$

*Choquet Integral*: Let $m$ be a fuzzy measure of Y, the Choquet Integral $C_m$ of $a = (a_1, \ldots, a_n)$. The criteria vector is defined by the equation:

$$C_m(a_1, \ldots, a_n) = \sum_{i=1}^{n}(a_i - a_{i-1})\, m(\{i, \ldots, n\}) \quad (5)$$

With $a_0 = 0$ and $a_1 \leq \ldots \leq a_n$.

In order to understand the concepts of the CI, we consider the following example where we try to merge three scores $s_1 = 0.7$, $s_2 = 0.8$ and $s_3 = 0.9$. The fuzzy measures associated to each score are: $m(s_1) = m^1 = 0.35$, $m(s_2) = m^2 = 0.25$, $m(s_3) = m^3 = 0.3$. We obtain the parameter $\lambda$ according to (4). The parameter $\lambda$ can be obtained by taking the unique root $\lambda > -1$, that is $\lambda = 0.361$. Following (3), we calculate the values of the fuzzy measurements on the subset of the scores as included in Table IV.

TABLE IV
THE VALUES OF THE FUZZY MEASUREMENT

| Subset | The fuzzy measurements |
|---|---|
| $\{s_1\}$ | $m^1 = 0.35$ |
| $\{s_2\}$ | $m^2 = 0.25$ |
| $\{s_3\}$ | $m^3 = 0.3$ |
| $\{s_1, s_2\}$ | $m^{12} = 0.631$ |
| $\{s_1, s_3\}$ | $m^{13} = 0.687$ |
| $\{s_3, s_2\}$ | $m^{23} = 0.577$ |
| $\{s_1, s_2, s_3\}$ | $m^{123} = 1$ |

To complete the calculation of final fusion score, we rearrange the scores and these yield values of the fuzzy measurements: $s_1 = 0.7 < s_2 = 0.8 < s_3 = 0.9$.

Using (5), the fusion value of the Choquet integral is: $C_m = (0.7 - 0)\, m(\{s1, s2, s3\}) + (0.8 - 0.7)\, m(\{s2, s3\}) + (0.9 - 0.8)\, m(\{s3\}) = 0.787$.

### B. The Genetic Algorithms

The Genetic Algorithms (GA) has been developed by Holland in 1975. It represents a stochastic optimization tool based on the mechanisms of natural selection and genetics. The GA operates with a population formed by a set of individuals called *chromosomes*. Every chromosome is constituted by a set of *genes*. Table V describes the principal parameters involved in a genetic algorithm [42].

TABLE V
THE PRINCIPAL PARAMETERS OF A GA

| | Description | Methods |
|---|---|---|
| Coding of chromosomes | This is the way that the chromosomes are represented. | • Real coding.<br>• Binary coding. |
| Initializing the population | This is the set of individuals, which constitutes the initial population. | • Random initialization.<br>• Initialization with existing solutions. |
| Fitness function | The fitness function associates a value to each individual. It can be either mono or multi criterion. | • No scaling.<br>• Linear scaling.<br>• Sigma truncation… |
| Parent selection | Select the best individuals of the current population to build new descendants | • Rank selection.<br>• Roulette wheel selection.<br>• Tournament selection<br>• Uniform selection… |
| Crossover | The crossovers used to form offspring with characteristics from parents where usually the best features are transmitted to the next generation. | • Arithmetic crossover<br>• BLX crossover.<br>• Linear crossover.<br>• Extended crossover … |
| Mutation | The mutation is to change or switch the values of genes on chromosome in order to constitute dissimilar individuals. | • Random mutation.<br>• Nonuniform mutation … |

In a problem of optimization by the genetic algorithms, the first step is to initialize the population randomly or with existing solutions. The second step involves a cost for each individual via a fitness function respecting the principle that the individuals survive well-adaptably. The third step is the reproduction: parents are selected by a method that favors the best of them; a crossover will give (new individuals) inheriting some of the characters of their parents. Finally, a mutation changes the value of some genes to prevent the establishment of a similar population unable to evolve [43].

### V. THE FUSION MODEL BASED ON THE CHOQUET INTEGRAL AND THE GA

The input of the fusion module is fed by three scores for the three considered modalities. By scores, we will refer to a measure of similarity that the identity of the candidate is supposed to be. The fusion approach that we propose is based on the CI with the fuzzy measurements generated by the GA as shown in Fig. 3.

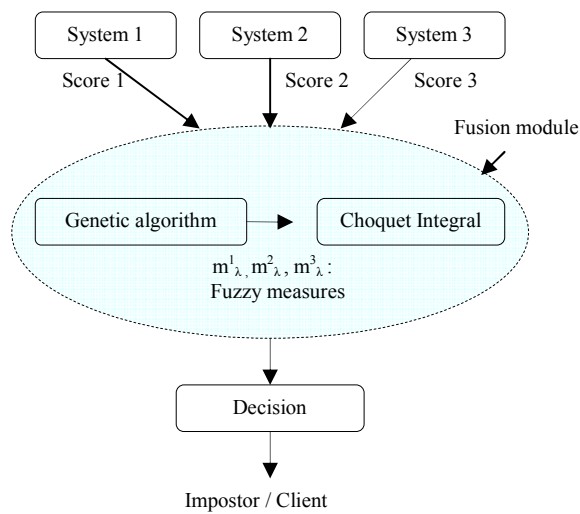

Fig. 3 Schema of the fusion of scores by the Choquet integral

The process of calculating the global score derived from the IC is given by the following algorithm:

**Step 1**: Initialization of the first generation P with solutions given by an expert election.
$P = (C_1, C_2, \ldots C_i, \ldots, C_N)$, where $C_i$ is the $i^{th}$ chromosome in

a population of *N* size.

**Step 2**: Calculation of the fitness function of the first population. Our fitness function is mono-criterion; it has as objective the minimization of Equal Error Rate during the classification. It is described by:

*Fitness = Minimise (EER)*

**Step 3**: Uniform selection of parents. We have randomly selected some $C_i$ individuals of the population *P*. The probability that an individual is selected is equal to *1/N*.

**Step 4**: Linear crossover of parents. We have crossed parents already selected to generate three descendants $h_i$, *i = 1, 2, 3* such as:

$$h_1 = 0.5\ (C_1 + C_2)$$
$$h_2 = (1.5\ C_1) - (0.5 C_2)$$
$$h_3 = (0.5\ C_1) + (1.5 C_2)$$

**Step 5**: Non-uniform mutation of descendants. We have applied a non-uniform mutation on the chromosomes output from the process of crossing. The mutation operator gives a Cm chromosome from an $h_i$ chromosome, such as:

$$C_m = h_i \pm y(1 - s^{(1 - \frac{itt}{gm})})$$

where *s*: a random number of the interval [0, 1].

*y*: the upper bound of the domain of variation of chromosome.

*itt*: value of the current iteration.

$g_m$: the maximum number of generation.

**Step 6**: Evaluation of the score of adaptation of a $P_{i+1}$ population. The descendants found in step 5 constitute the fuzzy measures for each unimodal system

**Step 6.1**: Calculation of the parameter λ (4).

**Step 6.2**: Calculation of the fuzzy measures of the subsets (3).

**Step 6.3**: Calculation of the score of the fusion of three unimodal systems by the CI (5).

**Step 6.4**: Calculation of the EER.

If the EER is less than a predefined threshold or the maximum number of iterations is reached, then the algorithm stops; if not, return to Step 3.

## VI. Experimentation and Results

The experiments have been performed on an Intel Dual-Core PC, having 1.73GHz, 1GB RAM, with the environment Matlab R2007 and Visual C++ under the Windows XP platform. To confirm the validity of the proposed fusion approach, we have implemented it on two different multimodal databases. The first database is relative only to synthetic scores. The second one is a biometric database relating to face, fingerprint and palmprint.

### A. Synthetic Database

The synthetic scores are derived from three virtual methods ($M_1$, $M_2$ and $M_3$) corresponding to 60 persons ($P_1$ to $P_{30}$: Clients scores, Table VI, $P_{60}$ to $P_{31}$: impostors' scores, Table VII). The scores are normalized between 0 and 1. It has been selected to cover all the combinations which may confront a fusion module of scores.

TABLE VI
THE SYNTHETIC CLIENT SCORES FROM THE VIRTUAL THREE MODALITIES

|  | $M_1$ | $M_2$ | $M_3$ |  | $M_1$ | $M_2$ | $M_3$ |
|---|---|---|---|---|---|---|---|
| $P_1$ | 0.98 | 0.98 | 0.98 | $P_{16}$ | 0.9 | 0.8 | 0.1 |
| $P_2$ | 0.98 | 0.98 | 0.6 | $P_{17}$ | 0.8 | 0.75 | 0.15 |
| $P_3$ | 0.98 | 0.6 | 0.98 | $P_{18}$ | 0.7 | 0.62 | 0.35 |
| $P_4$ | 0.98 | 0.6 | 0.6 | $P_{19}$ | 0.68 | 0.68 | 0.45 |
| $P_5$ | 0.98 | 0.7 | 0.6 | $P_{20}$ | 0.75 | 0.75 | 0.3 |
| $P_6$ | 0.9 | 0.8 | 0.7 | $P_{21}$ | 0.6 | 0.9 | 0.1 |
| $P_7$ | 0.8 | 0.8 | 0.8 | $P_{22}$ | 0.65 | 0.95 | 0.15 |
| $P_8$ | 0.7 | 0.9 | 0.9 | $P_{23}$ | 0.85 | 0.55 | 0.3 |
| $P_9$ | 0.7 | 0.7 | 0.9 | $P_{24}$ | 0.8 | 0.4 | 0.6 |
| $P_{10}$ | 0.7 | 0.9 | 0.7 | $P_{25}$ | 0.8 | 0.1 | 0.6 |
| $P_{11}$ | 0.6 | 0.6 | 0.6 | $P_{26}$ | 0.8 | 0.3 | 0.3 |
| $P_{12}$ | 0.6 | 0.7 | 0.95 | $P_{27}$ | 0.4 | 0.7 | 0.8 |
| $P_{13}$ | 0.6 | 0.95 | 0.7 | $P_{28}$ | 0.3 | 0.15 | 0.63 |
| $P_{14}$ | 0.55 | 0.55 | 0.55 | $P_{29}$ | 0.4 | 0.6 | 0.35 |
| $P_{15}$ | 0.9 | 0.8 | 0.4 | $P_{30}$ | 0.45 | 0.2 | 0.25 |

For each authentication session, three scores feed the fusion module based on the CI. An optimization module by the GA generates the most appropriate fuzzy measures for our fusion module. Fig. 4 shows the variation of fuzzy measures and the error rate from one generation to another. We observe that the search space of solutions is well scanned and that the optimal solution is reached after about 175 generations.

TABLE VII
THE SYNTHETIC IMPOSTOR SCORES FROM THE VIRTUAL THREE MODALITIES

|  | $M_1$ | $M_2$ | $M_3$ |  | $M_1$ | $M_2$ | $M_3$ |
|---|---|---|---|---|---|---|---|
| $P_{31}$ | 0.1 | 0.1 | 0.1 | $P_{46}$ | 0.4 | 0.2 | 0.75 |
| $P_{32}$ | 0.1 | 0.1 | 0.3 | $P_{47}$ | 0.3 | 0.1 | 0.55 |
| $P_{33}$ | 0.1 | 0.3 | 0.3 | $P_{48}$ | 0.2 | 0.05 | 0.65 |
| $P_{34}$ | 0.4 | 0.1 | 0.1 | $P_{49}$ | 0.15 | 0.1 | 0.55 |
| $P_{35}$ | 0.4 | 0.4 | 0.15 | $P_{50}$ | 0.15 | 0.1 | 0.7 |
| $P_{36}$ | 0.4 | 0.15 | 0.4 | $P_{51}$ | 0.15 | 0.4 | 0.8 |
| $P_{37}$ | 0.4 | 0.4 | 0.4 | $P_{52}$ | 0.35 | 0.7 | 0.1 |
| $P_{38}$ | 0.25 | 0.45 | 0.45 | $P_{53}$ | 0.35 | 0.55 | 0.3 |
| $P_{39}$ | 0.25 | 0.25 | 0.25 | $P_{54}$ | 0.15 | 0.65 | 0.2 |
| $P_{40}$ | 0.35 | 0.35 | 0.35 | $P_{55}$ | 0.15 | 0.55 | 0.4 |
| $P_{41}$ | 0.25 | 0.45 | 0.4 | $P_{56}$ | 0.15 | 0.55 | 0.6 |
| $P_{42}$ | 0.05 | 0.3 | 0.05 | $P_{57}$ | 0.6 | 0.3 | 0.15 |
| $P_{43}$ | 0.05 | 0.05 | 0.3 | $P_{58}$ | 0.6 | 0.55 | 0.15 |
| $P_{44}$ | 0.4 | 0.4 | 0.6 | $P_{59}$ | 0.6 | 0.15 | 0.55 |
| $P_{45}$ | 0.4 | 0.1 | 0.6 | $P_{60}$ | 0.6 | 0.55 | 0.55 |

The fuzzy measurements associated with the optimal solution are given in Table VIII. The recorded error rate is 5%.

TABLE VIII
THE VALUES OF THE FUZZY MEASUREMENT FOR THE OPTIMAL SOLUTION

| Subset | Fuzzy measures |
|---|---|
| $\{M_1\}$ | $m^1 = 0.411$ |
| $\{M_2\}$ | $m^2 = 0.547$ |
| $\{M_3\}$ | $m^3 = 0.362$ |
| $\{M_1, M_2\}$ | $m^{12} = 0.820$ |
| $\{M_1, M_3\}$ | $m^{13} = 0.682$ |
| $\{M_3, M_2\}$ | $m^{23} = 0.788$ |
| $\{M_1, M_2, M_3\}$ | $m^{123} = 1$ |

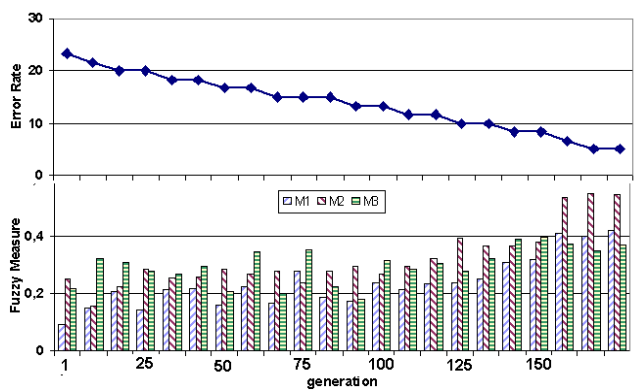

Fig. 4 Variation of the Fuzzy measurements and the error rate depending on the number of generation

Fig. 5 illustrates an overview of the distribution of client/impostor scores before and after the fusion. Before the fusion, we observe that the space of scores is divided into three areas: two extreme zones where the acceptance is certain and the rejection is absolute, an intermediate zone characterized by an overlap of the client/impostor scores. This overlap has been reduced after the fusion.

In order to demonstrate the contribution of our fusion approach, we have compared it with conventional fusion techniques (AND, OR, PROD, Mean, Majority Voting). The obtained performances are given in Table IX.

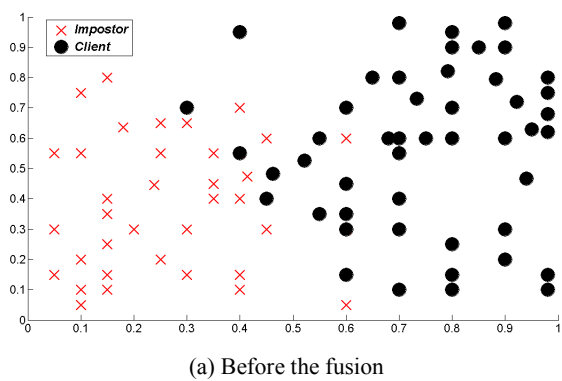

(a) Before the fusion

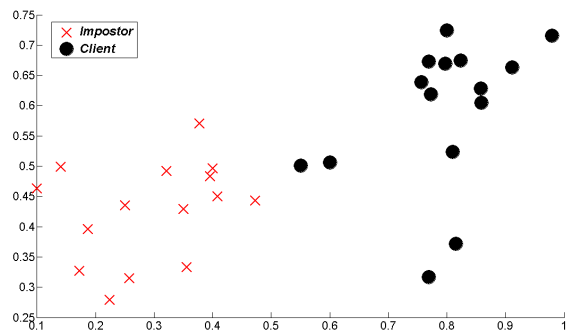

(b) After the fusion

Fig. 5 Distribution of the impostor scores and the client scores before and after the fusion by the CI

TABLE IX
COMPARISON OF THE OBTAINED PERFORMANCES FOR THE PROPOSED METHOD AND OTHER TECHNIQUES KNOWN IN THE LITERATURE

|  |  | Error Rate (%) |
|---|---|---|
| **Modality 1 :** |  | 13.33 |
| **Modality 2 :** |  | 20 |
| **Modality 3 :** |  | 38.33 |
| **Classical fusions techniques** | AND | 28.33 |
|  | OR | 30 |
|  | PROD | 40 |
|  | Mean | 10.33 |
|  | Vote | 13.33 |
| **Integral Choquet fusion** |  | 5 |

*B. Multimodal Biometric Database*

*Face database*: The face images are obtained from the face94 database of the University of Essex. The face database consists of 153 subjects with 20 face images available for each subject. The subjects sit at a fixed distance from the camera and are asked to speak. The speech is used to introduce facial expression variation. All face images resolution are RGB images, 180 × 200 pixels in JPEG format [41]. Fig. 6 shows face image samples of 10 users.

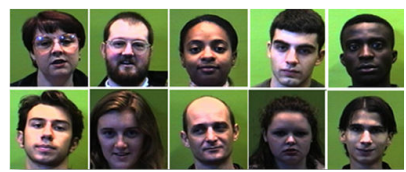

Fig. 6 Face image samples

*Palmprint database:* The palmprint images are obtained from the Hong Kong Polytechnic University 2D_3D palmprint database. The database consists of 400 subjects with 20 palmprint images available for each subject. All palmprint are greyscale images, 128 × 128 resolution which contain the ROI of the palmprint of the right hand [39]. Fig. 7 shows palm print image samples of 10 users.

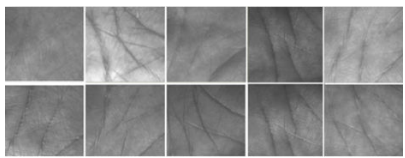

Fig. 7 Palmprint image samples

*Fingerprint database*: The fingerprint images are obtained from the Hong Kong Polytechnic University HRF database. The HFR database contains 1480 fingerprint images from 148 fingers. All fingerprints are greyscale images, 640 × 480 resolution [37]. Fig. 8 shows fingerprint image samples of 10 users.

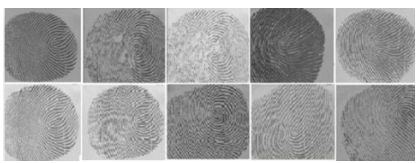

Fig. 8 Fingerprint image samples

The fusion of the three biometric systems is made at the score level by the CI. An optimization module by the AG calculates the most appropriate fuzzy measures for our fusion module. Fig. 9 shows the variation of the EER from one generation to another. We observe the same results for those obtained in the case of the synthetic database. The optimal solution is reached after 960 generations. The recorded EER is 0.46%.

Fig. 9 illustrates an overview of the distribution of client/impostor scores before and after the fusion. Before the fusion, we see a significant overlap particularly in the interval [0.5, 0.7]. This overlap between the two classes is directly responsible for errors in classifying unimodal systems. We find in Fig. 10 (b) that our fusion approach through the CI and the GA has limited the overlap between the two classes, which has enabled improving the performance of the multimodal system.

In order to demonstrate the contribution of our fusion approach, we have compared it with the conventional fusion techniques (AND, OR, PROD, Mean, Majority Voting). Fig. 11 gives the recorded results.

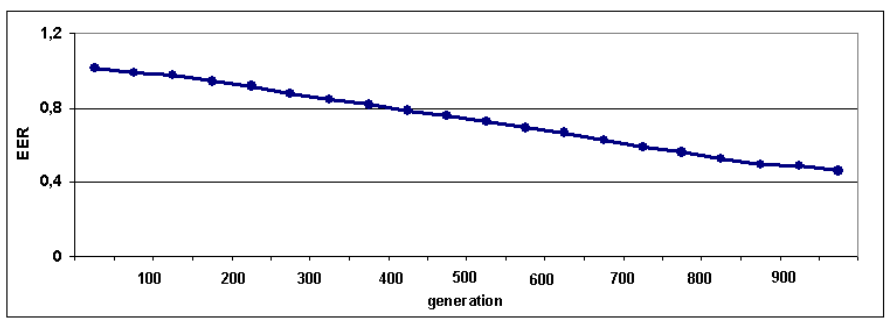

Fig. 9 The variation of the EER from one generation to another

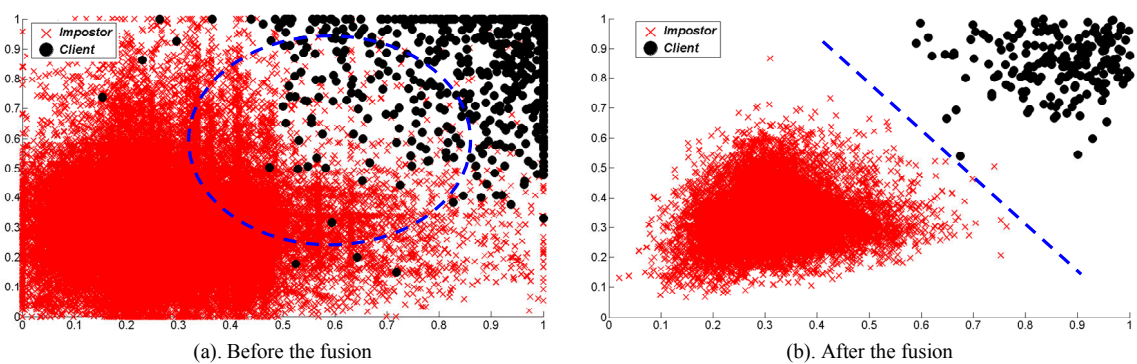

(a). Before the fusion     (b). After the fusion

Fig. 10 Distribution of the impostor scores and the client scores before and after the fusion by the CI

The analysis of the recorded results shows that the fusion of the three unimodal systems has improved significantly the performance of the multimodal system. Indeed, the EER has increased from 2.51% (best unimodal system) to 0.46% (best multimodal system). As we can expect, the 'AND/OR' fusion techniques based on a combination of binary decision give results that are not interesting, hence finding approximately the performance of the best unimodal system (Face). In contrast, the fusion approaches at the score level provide good results; especially, the fusion approach by the CI outperforms the conventional fusion techniques.

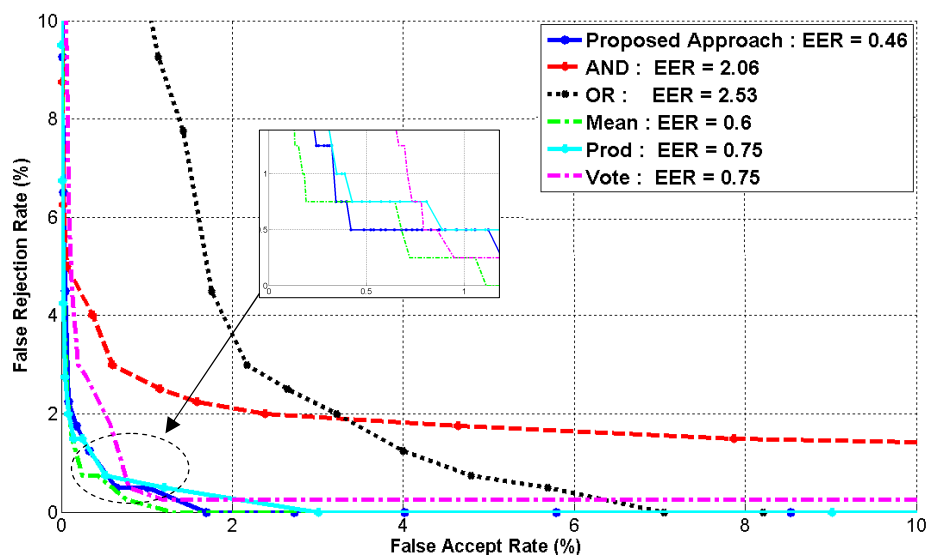

Fig. 11 ROC curves of different fusion rules

## VII. CONCLUSION

In this paper, we have proposed three biometric systems based on face, fingerprint and palmprint. In the three systems, the characterization is based on an analysis of texture by the wavelet transformation; the classification is assured by a modular architecture at the base of support vector machines. We have also proposed an introduction of the fusion of the three unimodal systems to level the scores by the Choquet integral. We have shown that the fuzzy measures calculated by the genetic algorithms contribute to improve the multimodal system performance.

## REFERENCES

[1] Ben Khalifa, A. Ben Amara, N.E. "Exploration of the Choquet integral for the fusion of biometric modalities," IEEE International Multi-Conference on Systems, Signals and Devices (SSD), 2012, pp. 1 – 6

[2] Ben Khalifa, A. Ben Amara, N.E. "Bimodal biometric verification with different fusion levels," IEEE International Multi-Conference on Systems, Signals and Devices (SSD), 2009, pp. 1 – 6.

[3] Ben Khalifa, A. Ben Amara, N.E. "Fusion at the feature level for person verification based on off line handwriting and signature," IEEE International Conference on Signals, Circuits and Systems, 2008, pp. 1 – 5.

[4] P. Verlinde, P. Druyts, G. Chollet, and M. Acheroy. "A multi-level data fusion approach for gradually upgrading the performances of identity verifcation systems," In Sensor Fusion: Architectures, Algorithms, and Applications III, volume 3719, Orlando, USA, April 1999.

[5] Ferrer, Miguel A.; Morales, Aythami; Travieso, Carlos M.; Alonso, Jesws B.; "Low Cost Multimodal Biometric identification System Based on Hand Geometry, Palm and Finger Print Texture," IEEE International Carnahan Conference on Security Technology, 2007. pp.52 – 58.

[6] Israel, S.A.; Scruggs, W.T.; Worek, W.J.; Irvine, J.M.; "Fusing face and ECG for personal identification," Proceedings Applied Imagery Pattern Recognition Workshop, 2003. pp.226 – 231.

[7] Rattani, Ajita; Kisku, D. R.; Bicego, Manuele; Tistarelli, Massimo; "Robust Feature-Level Multibiometric Classification," Biometric Consortium Conference, 2006. pp.1 – 6

[8] Z. Liu and S. Sarkar, "Outdoor recognition at a distance by fusing gait and face," Image Vision Comput., vol. 25, pp. 817-832, 2007.

[9] Jinfeng Yang, Xu Zhang, "Feature-level fusion of fingerprint and finger-vein for personal identification," Pattern Recognition Letters, Volume 33, Issue 5, 1 April 2012, pp. 623-628.

[10] Carlos M. Travieso, Jianguo Zhang, Paul Miller, Jesús B. Alonso, Miguel A. Ferrer, "Bimodal biometric verification based on face and lips," Neurocomputing, Volume 74, Issues 14–15, July 2011, pp. 2407-2410.

[11] Abdallah Meraoumia, Salim Chitroub, Ahmed Bouridane, "Fusion of Finger-Knuckle-Print and Palmprint for an Efficient Multi-biometric System of Person Recognition," IEEE International Conference on Communications (ICC), 2011, pp. 1 – 5.

[12] S. Pigeon and L. Vandendorpe. "Multiple experts for robust face authentication," In Optical Security and Counterfeit Deterrence Techniques II, pp. 166–177, California, January 1998. Proceedings of SPIE no 3314.

[13] Jingyan Wang; Yongping Li; Xinyu Ao; Chao Wang; Juan Zhou, "Multi-modal biometric authentication fusing iris and palmprint based on GMM," IEEE Workshop on Statistical Signal Processing, 2009. pp. 349 – 352.

[14] Sheetal Chaudhary, Rajender Nath, "A Multimodal Biometric Recognition System Based on Fusion of Palmprint, Fingerprint and Face," International Conference on Advances in Recent Technologies in Communication and Computing, 2009, pp. 596-600.

[15] Choquet G., Théorie des capacités. 1953.

[16] Grabisch M., "A new algorithm for identifying fuzzy measures and its application to pattern recognition," ICFS 1995, pp. 145-150.

[17] Sugeno M., "Fuzzy measures and fuzzy integrals – A survey," FADP 1977, pp. 89-102.

[18] Kwak K., Pedrycz W., "Face recognition: A study in information fusion using fuzzy integral," Pattern Recognition Letters 26 2005, pp. 719-733.

[19] Andrey Temkoa, Dušan Machob, Climent Nadeua, "Fuzzy integral based information fusion for classification of highly confusable non-speech sounds," Pattern Recognition 41, 2008, pp. 1814 – 1823.

[20] Hirota, K.; Vu, H.A.; Le, P.Q.; Fatichah, C.; Liu, Z.; Tang, Y.; Tangel, M.L.; Mu, Z.; Sun, B.; Yan, F.; Masano, D.; Thet, O.; Yamaguchi, M.; Dong, F.; Yamazaki, Y. "Multimodal gesture recognition based on Choquet integral," IEEE International Conference on Fuzzy Systems, 2011, pp. 772 – 776.

[21] Belahcene, M.; Ouamane, A.; Ahmed, A.T. "Fusion by combination of scores multi-biometric systems," European Workshop on Visual Information Processing (EUVIP), 2011, pp. 252 – 257.

[22] Jullien, S.; Valet, L.; Mauris, G.; Bolon, P.; Teyssier, S. "An Attribute Fusion System Based on the Choquet Integral to Evaluate the Quality of Composite Parts," IEEE Transactions on Instrumentation and Measurement, Volume: 57, Issue: 4, 2008, pp. 755 - 762.

[23] Wen-Chih Lin; Chih-Sheng Huang; Jeng-Ming Yih; Der-Bang Wu; Yen-Kuei Yu, "Multiple SVM classification syatem based on Choquet integral with respect to composed measure of L-measure and Delta-measure," International Conference on Machine Learning and Cybernetics (ICMLC), 2010, pp. 2396- 2401.

[24] Kuan-Kai Huang; Jiunn-I Shieh; Kuei-Jen Lee; Shih-Neng Wu, "Applying a generalized choquet integral with signed fuzzy measure based on the complexity to evaluate the overall satisfaction of the

patients," International Conference on Machine Learning and Cybernetics (ICMLC), 2010, pp. 2377- 2382.
[25] Afef Denguir Rekik, Un Cadre Possibiliste pour l'Aide à la Decision Multicritère et Multi-acteurs Application au Marketing et au Benchmarking de sites E-commerce, Ph.D. thesis, Université de Savoie (2007).
[26] Asma Melki, 'Système d'aide à la régulation et évaluation des transports multimodaux intégrants les cybercars', Ph.D. thesis, Ecole Centrale de Lille (2008).
[27] Mohamed, M.A.; Abou-Elsoud, M.E.; Eid, M.M., "Automated face recogntion system: Multi-input databases," International Conference on Computer Engineering & Systems (ICCES), 2011. pp. 273 – 280.
[28] Rahman, S.; Naim, S.M.; Al Farooq, A.; Islam, M.M., "Curvelet texture based face recognition using Principal Component Analysis," International Conference on Computer and Information Technology (ICCIT), 2010. pp. 45 - 50.
[29] Neo, H.F.; Teo, C.C.; Teoh, A.B.J., "Development of Partial Face Recognition Framework," International Conference on Computer Graphics, Imaging and Visualization (CGIV), 2010, pp. 142 - 146.
[30] Neo Han Foon; Ying-Han Pang; Jin, A.T.B.; Ling, D.N.C., "An efficient method for human face recognition using wavelet transform and Zernike moments," International Conference on Computer Graphics, Imaging and Visualization, 2004. pp. 65 - 69.
[31] Nicholl, P.; Bouchaffra, D.; Amira, A.; Perrott, R.H., "Multiresolution Hybrid Approaches for Automated Face Recognition," Conference on Adaptive Hardware and Systems, 2007. pp. 89 - 96.
[32] Jones, C.; Abbott, A.L., "Color face recognition by hypercomplex Gabor analysis," International Conference on Automatic Face and Gesture Recognition, 2006. pp. 1 – 6.
[33] Yinghua Lu; Yao Fu; Jinsong Li; Xiaolu Li; Jun Kong, "A Multi-modal Authentication Method Based on Human Face and Palmprint," International Conference on Future Generation Communication and Networking, 2008. pp. 193- 196.
[34] Wen-Ying Ma; Sheng Li; Yong-Fang Yao; Chao Lan; Shi-Qiang Gao; Hui Tang; Xiao-Yuan Jing, "Multi-Modal Biometrics Pixel Level Fusion and KPCA-RBF Feature Classification for Single Sample Recognition Problem," International Congress on Image and Signal Processing, 2009. pp. 1 – 5.
[35] Yong Jian Chin; Thian Song Ong; Goh, M.K.O., Bee Yan Hiew, "Integrating Palmprint and Fingerprint for Identity Verification," International Conference on Network and System Security, 2009. pp. 437- 442.
[36] Kekre, H.B.; Tanuja; Sarode, K.; Tirodkar, A.A, "A study of the efficacy of using Wavelet Transforms for Palm Print Recognition," International Conference on Computing, Communication and Applications (ICCCA), 2012, pp. 1 - 6.
[37] Feng Liu, Qijun Zhao, Lei Zhang, and David Zhang, "Fingerprint Pore Matching based on Sparse Representation," International Conference on Pattern Recognition (ICPR'10), 2010, pp. 1630 – 1633.
[38] Qijun Zhao, Lei Zhang, David Zhang, and Nan Luo, "Direct Pore Matching for Fingerprint Recognition," International Conference on Biometrics, 2009, pp. 597-606.
[39] David Zhang, Wai-Kin Kong, Jane You and Michael Wong, "On-line palmprint identification," IEEE Transactions on Pattern Analysis and Machine Intelligence, pp. 1041-1050, 2003.
[40] Jain, A.K.; Ross, A.; Prabhakar, S, "An introduction to biometric recognition," IEEE Transactions on Circuits and Systems for Video Technology, 2004, pp. 4 - 20.
[41] D. Hond, L. Spacek "Distinctive Descriptions for Face Processing," Proceedings of the 8th British Machine Vision Conference BMVC97, Colchester, England, pp. 320-329,1997.
[42] Holland J.H. "Adaptation in natural and artificial systems," MIT Press 1975.
[43] Man K.F., Tang K.S., Kwong S. "Genetic algorithms Concepts and designs," 2000, Springer.